\documentclass{article}
\usepackage{ijcai17}

\usepackage{times}
\usepackage{array}
\usepackage{multirow}
\usepackage{slashbox} 
\usepackage{subfig}
\usepackage{tikz}
\usepackage{indentfirst}
\usepackage{amsmath}
\usepackage{algorithm}
\usepackage{algorithmic}

\title{Conceptualization Topic Modeling\thanks{Submission to IJCAI 2017}}
\author{
Yi-Kun Tang$^\spadesuit$, Xian-Ling Mao$^\spadesuit$\thanks{\hspace{0.15cm}Corresponding author.}, Heyan Huang$^\spadesuit$, Guihua Wen$^\clubsuit$\\
 $^\spadesuit$Department of Computer Science, Beijing Institute of Technology, China \\ 
  $^\clubsuit$Department of Computer Science and Technology, South China University of Technology, China\\ 
  {\tt \{tangyk, maoxl, hhy63\}@bit.edu.cn} ,  {\tt crghwen@scut.edu.cn}
}
\begin{document}

\maketitle

\begin{abstract}
Recently, topic modeling has been widely used to discover the abstract topics in text corpora.   
Most of the existing topic models are based on the assumption of   three-layer hierarchical Bayesian structure, i.e. each document is modeled as a probability distribution over topics, and each topic is a probability distribution over words.  
However, the assumption is not optimal.  
Intuitively, it's more reasonable to assume that each topic is a probability distribution over concepts, and then each concept is a probability distribution over words, i.e. adding a latent concept layer between topic layer and word layer in traditional three-layer assumption.  
In this paper, we verify the proposed assumption by incorporating the new assumption in two representative topic models, and obtain two novel topic models.   
Extensive experiments were conducted among the proposed models and corresponding baselines, and the results show that the proposed models significantly outperform the baselines in terms of case study and perplexity, which means the new assumption is more reasonable than traditional one.  
\end{abstract}

\section{Introduction}
In recent years, topic modeling is becoming more and more popular in identifying latent semantic components in text corpora. Lots of topic models have been proposed. The existing topic models can be divided into four categories: \textit{Unsupervised non-hierarchical topic models}   \cite{deerwester1990indexing,hofmann1999probabilistic,blei2003latent,yao2016concept},  \textit{Unsupervised hierarchical topic models}     \cite{blei2003hierarchical,teh2006hierarchical,joshi2016political}, and their corresponding supervised counterparts   \cite{ramage2009labeled,mao2012sshlda:,magnusson2016dolda}.   

The basic assumption of most existing topic models is that each document is modeled as a probability distribution over topics, and each topic is directly a probability distribution over words, i.e.  three-layer hierarchical Bayesian structure, shown in Figure 1 (a).   


However, this assumption is not optimal, because it does not consider the importance of the concepts in topics.   
Concepts are very important in natural language and textual semantic understanding. Concepts can also help people better understand knowledge, as psychologist Gregory Murphy wrote: \emph{"Concepts are the glue that holds our mental world together"} \cite{murphy2004the}.  

\begin{figure}[tb]
\centering 
\includegraphics[width=0.48\textwidth]{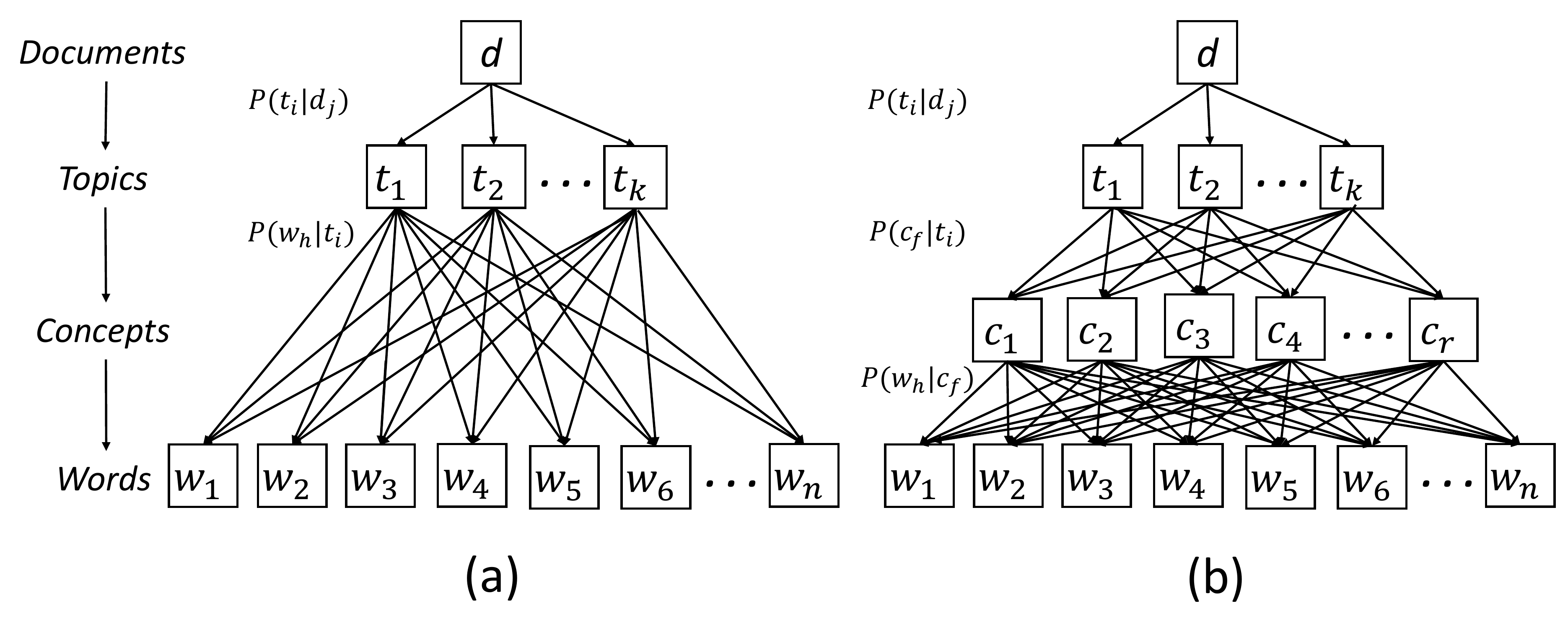}
\caption{(a) Three-layer hierarchical Bayesian structure of existing topic models; (b) Four-layer hierarchical Bayesian structure of conceptualization topic modeling.}  
\end{figure}


Intuitively, it's more reasonable that if  we add a latent concept layer between topic layer and word layer in traditional three-layer assumption, i.e. a four-layer hierarchical Bayesian structure, shown in Figure 1 (b). In this novel assumption, each document is considered as a probability distribution over topics, each topic is a probability distribution over concepts, and each concept is a probability distribution over words.  
The assumption is similar to the writing process.  
For example, if we want to write a article about the topic ``\emph{military}'', we then focus on the concepts related to the topic, such as \emph{army}, \emph{navy} and \emph{air force}. Finally, we select related words from these concepts, maybe the word \emph{tank} from the concept \emph{army}, the word \emph{torpedo} from the concept \emph{navy}, and the word \emph{fighter} from the concept \emph{air force}.  

As we known, Latent Dirichlet Allocation (LDA) \cite{blei2003latent} is the beginning of topic modeling, and is the most important component in all kinds of topic models. 
If the novel assumption performs better than the traditional one in LDA, it's reasonable to infer that the novel assumption is more suitable for topic modeling than the traditional one.  
Thus, in this paper, we first propose a novel topic model, called  Conceptualization Latent Dirichlet Allocation (CLDA), which applies the novel four-layer assumption in LDA, to verify our assumption. Furthermore, we also apply the novel assumption in a supervised topic model, Labeled LDA (LLDA) \cite{ramage2009labeled}, to proof the novel assumption is more effective.   
The distribution of each concept over words in our models can be obtained from Probase knowledge base \cite{wu2012probase:}, which is a universal probabilistic taxonomy
concept knowledge base.  

  
The rest of the paper is organized as follows. In Section 2, we review the related work. 
In Section 3, two novel topic models, CLDA and CLLDA, are proposed by using new four-layer assumption. Extensive experiments on two real datasets are introduced in Section 4. Finally, we conclude the paper in      Section 5.  
\section{Related Work}

\subsection{Topic Modeling}
\label{sec:topic-modeling}

The existing topic models can be divided into four categories: \textit{Unsupervised non-hierarchical topic models, Unsupervised hierarchical topic models}, and their corresponding supervised counterparts.

Unsupervised non-hierarchical topic models are widely studied, such as LDA \cite{blei2003latent}, Probase-LDA \cite{yao2015incorporating} , TCC \cite{jayabharathy2014correlated} and COT \cite{yao2016concept} etc. The most famous one is Latent Dirichlet Allocation (LDA). LDA is similar to pLSA \cite{hofmann1999probabilistic}, except that in LDA the topic distribution is assumed to have a Dirichlet prior.   

However, the above models cannot capture the relation between super and sub topics. To address this problem, many models have been proposed to model the relations, such as Hierarchical LDA (HLDA) \cite{blei2004hierarchical}, Hierarchical Dirichlet processes (HDP) \cite{teh2006hierarchical}, Hierarchical PAM (HPAM) \cite{mimno2007mixtures}, PIE \cite{joshi2016political} and Guided HTM \cite{shin2016guided} etc.  The relations are usually in the form of a hierarchy, such as the tree or Directed Acyclic Graph (DAG).   

Although unsupervised topic models are sufficiently expressive to model multiple topics per document, they are inappropriate for labeled corpora because they are unable to incorporate the observed labels into their learning procedure.  Several modifications of LDA to incorporate supervision have been proposed in the literature, such as Supervised LDA \cite{blei2007supervised,blei2010supervised}, Prior-LDA  \cite{rubin2011statistical}, Partially LDA (PLDA) \cite{ramage2011partially}, NTM \cite{cao2015novel} and DOLDA \cite{magnusson2016dolda} etc.    

None of these non-hierarchical supervised models, however, leverage on dependency structure, such as parent-child relation, in the label space. 
 Lots of models, such as hLLDA \cite{petinot2011hierarchical}, HSLDA \cite{perotte2011hierarchically}, SSHLLDA \cite{mao2012sshlda:}, SHDP \cite{zhang2013supervised} and EHLLDA \cite{mao2015ehllda},  have been   proposed to solve the problem.

All of these topic models are mainly based on the assumption of   three-layer hierarchical Bayesian structure.  
However, the assumption is not optimal.  
Intuitively, it's more reasonable to add a latent concept layer between topic layer and word layer in traditional three-layer assumption.    
In this paper, we will verify the proposed assumption by incorporating the new assumption in two representative topic models.   

\subsection{Concept Knowledge Base}
It is easy for mankind to acquire the meaning of an article and extract the topics of the article, 
because there is a certain background conceptualized knowledge base in a brain.  
For example, when seeing a sentence: "Microsoft announced a project named, Microsoft Azure Information Protection.",  
a man will never mistake \emph{Microsoft} as a person or other things, because we have known that \emph{Microsoft} is a concept about software  company.  

However, machines cannot conceptualize what they read, which is a great challenge for  machines to understand natural language.  
Concept knowledge base is a kind of knowledge base that uses taxonomies and ontologies to obtain concepts and extract the relationships between instances and concepts. Therefore, concept knowledge base is a kind of tool to make machines understand nature language.

There are many existing concept knowledge bases, such as Probase  \cite{wang2015an,wu2012probase:}, Freebase and WordNet etc. Among them, Probase is a state-of-the-art one, which contains above 5.4 million concepts that is greater than other   concept knowledge bases. The main advantage of Probase is that it is the first to measure the correlation between instances and concepts with probabilities, while other concept knowledge bases use a boolean variable to represent relationships between instances and concepts.

Therefore, in this paper, we use Probase API \cite{wang2015an} to get the probability distribution of each concept over words.

\section{Conceptualization Topic Modeling}  
In this section, we will demonstrate that how to incorporate the four-layer assumption in unsupervised and supervised topic models, to verify the effectiveness of the novel assumption. For unsupervised topic modeling, we choose LDA as the manipulating object because it is the basic component of most existing topic models. For supervised topic modeling, we choose Labeled LDA \cite{ramage2009labeled} because it is one of the most representative supervised models.  

\subsection{Conceptualization LDA}
\label{sec:conc-lda}
To incorporate the four-layer assumption in LDA, we propose a novel topic model, called Conceptualization LDA (CLDA).   
 It models each document as a mixture of underlying topics. Different from existing topic models, CLDA assumes that each topic is a distribution over concepts rather than directly over words, and regards concepts as distributions over words.  

In addition, as for neologisms, which do not in the dictionary of the concept knowledge base, they will be regarded as new concepts. In other words, we define these neologisms as atomic concepts. In CLDA,   
the distribution of a concept over words is acquired from the concept knowledge base, Probase. The graphical model of CLDA is shown in Figure 2.  

\begin{figure}[htb!]
\centering
\includegraphics[width=0.45\textwidth]{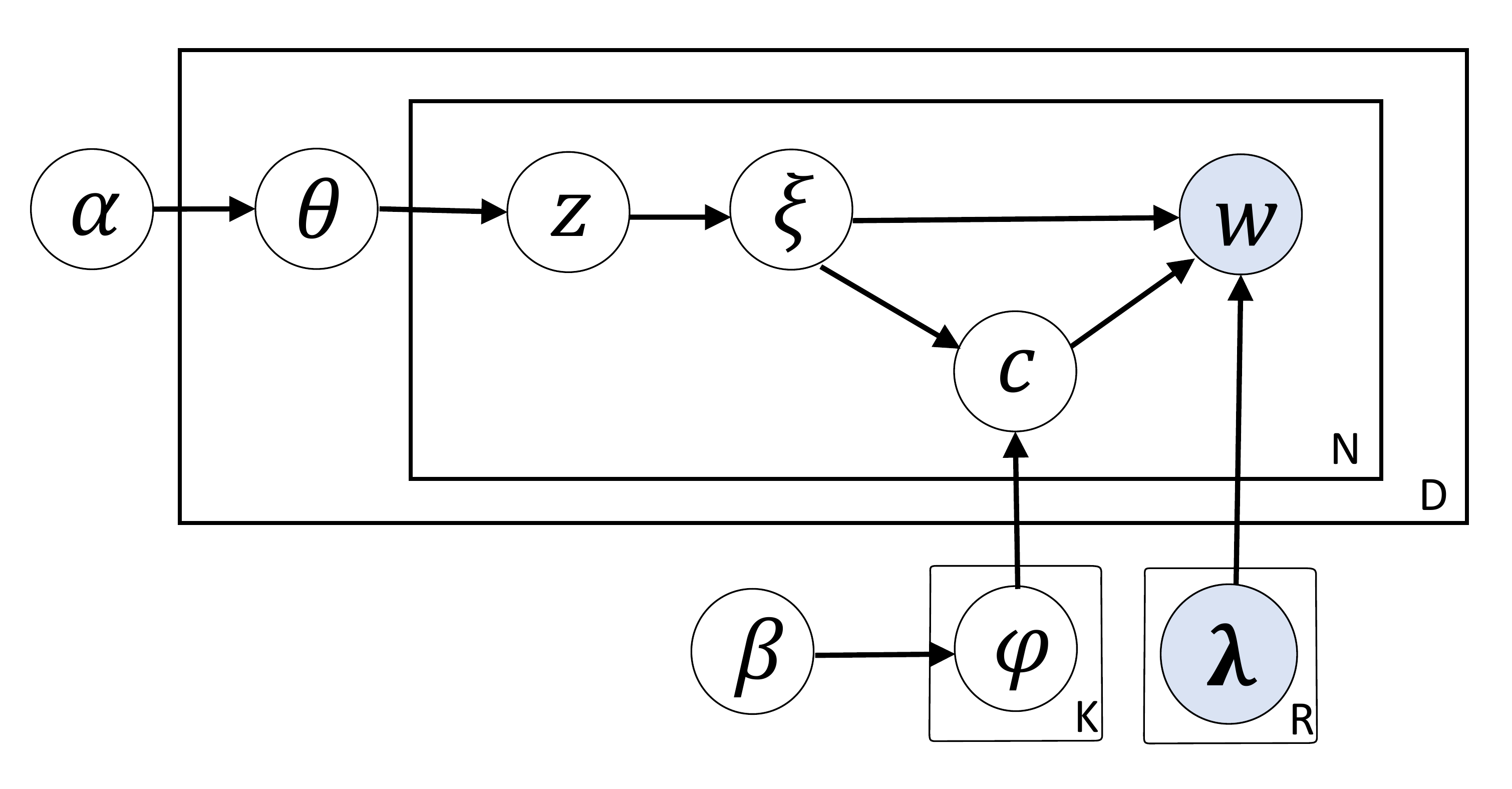}
\caption{Graphical model for CLDA. }
\end{figure}

In CLDA, each document consists of a group of words represented as $w^{(d)}=(w_{1},...,w_{N_{d}})$. $\alpha$ is the parameter of the Dirichlet distribution of the topic prior, and $\theta^{(d)}$ is the parameter of the multinomial distribution of the $d^{th}$ document. $\beta$ is the parameter of the Dirichlet distribution of the concept prior, and $\phi_{k}$ is the parameter of the multinomial distribution of the $k^{th}$ topic. $\lambda$ is the concept distribution over words gets from Probase. $m$ is the number of words that do not belong to any concept in the concept knowledge base, and R is the size of concept set. $z_{d,i}$ is the latent topic for the concept or atom concept of the $i^{th}$ word in the $d^{th}$ document. 
 $c_{d,i}$ is the concept of the $i^{th}$ word in the $d^{th}$ document.

The generative process of our CLDA is summarized in Algorithm 1. The generative process can be divided into three parts. Firstly, draw the concept and atom concept distribution from Dirichlet distribution for each topic in the datasets (line 1 $\sim$ 2). Secondly, draw the topic distribution for each document from Dirichlet distribution (line 3 $\sim$ 4). Finally, to generate the word $w_{d,i}$, we first select a latent topic $z_{i}$ (line 5 $\sim$ 6), and then generate a variable $\xi$ from Bernoulli distribution, where 0 indicates the word does not belong to any concept, and 1 indicates the word $w_{d,i}$ belongs to some concepts in the given concept knowledge base. If $\xi$ equals to 0, then generate a word from $Mult(\cdot|\phi_{z_{i}})$; otherwise, generate a concept from $Mult(\cdot|\phi_{z_{i}})$, and then select a word from the concept, which conditionally is related to the concept distribution from the knowledge base (line 7 $\sim$ 12).  

%
\subsubsection{Learning and Inference}
In this section, we use collapsed Gibbs sampling to estimate parameters. 

Specifically, if the word $w_{d,i}$ belongs to some concepts in the given concept knowledge base, the sampling probability for a topic and a concept in position $i$ in document $d$ can be expressed as follows:
\begin{equation}
\small
\begin{aligned}
&P(z_{d,i}=k,c_{d,i}=j|\textbf{w},\textbf{e}_{-(d,i)},\textbf{z}_{-(d,i)};\alpha,\beta) \\
\propto
&\frac{\beta_{k,c_{d,i}}+n^{(c_{d,i})}_{-(d,i),k}}{\sum^{E}_{x=1}\beta_{k,x}+n^{(\cdot)}_{-(d,i),k}}\cdot
\frac{\alpha_{k}+n^{(d)}_{-(d,i),k}}{\sum^{K}_{t=1}\alpha_{t}+n^{(d)}_{-(d,i),\cdot}}\cdot P(w_{d,i}|c_{d,i})
\end{aligned}
\end{equation}

\begin{algorithm}[tb]
\caption{ Generative process for CLDA. }
\begin{algorithmic}[1]
\STATE For each topic k $\in \{1,...,K\}$:
\STATE ~~~~Generate $\phi_{k}=(\phi_{k,1},...,\phi_{k,C},\phi_{k,C+1},...,\phi_{k,C+m})^{T}$ \\ $\sim Dir(\cdot|\beta)$
\STATE For each document d $\in \{1,...,D\}$:
\STATE ~~~~Generate $\theta^{(d)}=(\theta_{1},...,\theta_{k})^{T}\sim Dir(\cdot|\alpha)$
\STATE ~~~~For each i in \{1,...,$N_{d}$\}:
\STATE ~~~~~~~~Generate $z_{d,i}\in \{1,...,K\}\sim Mult(\cdot|\theta^{(d)})$
\STATE ~~~~~~~~Generate $\xi \sim Bernoulli$, where 0 indicates the word $w_{d,i}$ is an atom concept, and 1 indicates the word $w_{d,i}$ belongs to some concepts in the given concept knowledge base.
\STATE ~~~~~~~~If $\xi=0$:
\STATE ~~~~~~~~~~~~Generate $w_{d,i}\in\{1,...,V\}\sim Mult(\cdot|\phi_{z_{i}})$
\STATE ~~~~~~~~Else:
\STATE ~~~~~~~~~~~~Generate $c_{d,i}\in\{1,...,R\}\sim Mult(\cdot|\phi_{z_{i}})$
\STATE ~~~~~~~~~~~~~Select a word $w_{d,i}$ from $\lambda$, a probability distribution gets from Probase.
\end{algorithmic}
\end{algorithm}

And if the word $w_{d,i}$ does not belong to any concept in the given concept knowledge base, the sampling probability for a topic in position $i$ in document $d$ can be expressed as follows:
\begin{equation}
\small
\begin{aligned}
&P(z_{(d,i)}=k|\textbf{w},\textbf{e}_{-(d,i)},\textbf{z}_{-(d,i)};\alpha,\beta) \\
\propto
&\frac{\beta_{k,e_{d,i}}+n^{(e_{d,i})}_{-(d,i),k}}{\sum^{E}_{x=1}\beta_{k,x}+n^{(\cdot)}_{-(d,i),k}}\cdot
\frac{\alpha_{k}+n^{(d)}_{-(d,i),k}}{\sum^{K}_{t=1}\alpha_{t}+n^{(d)}_{-(d,i),\cdot}}
\end{aligned}
\end{equation}
where $\textbf{e}$ is the vector of concepts and atomic concepts related to the words. $e$ denotes a concept or an atomic concept. $E$ is the number of concepts and atomic concepts. $n^{(\cdot)}_{-(d,i),k}$ is the count of concepts and atomic concepts in $\textbf{e}$ in topic k without $z_{d,i}$. $n^{(d)}_{-(d,i),k}$ is the number of tokens in $\textbf{e}$ assigned to topic k in document d without $z_{d,i}$, and $n^{(d_{i})}_{-(d,i),\cdot}$ indicates a summation over that dimension.
In Eq. (1), $n^{(c_{i})}_{-i,k}$ is the count of concept $c_{i}$ in topic k, that does not include the current assignment $z_{(d,i)}$. And the conditional probability $P(w_{d,i}|c_{d,i})$ describes the probability of word $w_{d,i}$ in concept $c_{d,i}$, which can be obtained from Probase. In Eq. (2), $n^{(e_{i})}_{-(d,i),k}$ is the count of atom concept $e_{d,i}$, non-concept word, in topic k, that does not include the current assignment $z_{d,i}$.

Finally, the parameters can be estimated as follows:  

\begin{equation}
\hat{\phi}_{e,k}=
\frac{\beta_{k,e}+n^{(e)}_{\cdot,k}}{\sum^{E}_{x=1}\beta_{k,x}+n^{(\cdot)}_{\cdot,k}}
\end{equation}

\begin{equation}
\begin{aligned}
\hat{\theta}_{k}^{(d)}=\frac{\alpha_{k}+n^{(d)}_{\cdot,k}}{\sum^{K}_{t=1}\alpha_{t}+n^{(d)}_{\cdot,\cdot}}
\end{aligned}
\end{equation}
where $e$ denotes a concept or an atomic concept.  
The two equations for parameter estimation are important. We can use the topic-specific distribution $\phi$ to obtain topical abstracts for topics; meanwhile the topic distribution for each document $\theta$ can be used to discover the most relevant topics for a document and find documents with similar topics.

\subsection{Conceptualization Labeled LDA}
  \label{sec:conc-llda-sect}  
The proposed four-layer Bayesian assumption can be used in most of existing topic models, and we have demonstrated that the assumption can be used in unsupervised topic model, i.e. LDA. In this section, we will further demonstrate the use of the assumption in supervised topic modeling.    
Labeled Latent Dirichlet Allocation (Labeled LDA) \cite{ramage2009labeled} which is a classical supervised topic model, will be extended by incorporating conceptualization assumption. The novel model is called Conceptualization Labeled Latent Dirichlet Allocation (CLLDA).  

Labeled LDA is very similar to LDA.  Different with LDA, Labeled LDA assumes that the topics of each document  are restricted to its labels. The topic distribution of each document in  Labeled LDA is generated from a Dirichlet distribution, whose dimensionality of the prior parameter is the same as the number of labels of each document, rather than the number of the total topics of the datasets in LDA.  Thus, CLLDA is also similar to CLDA.  


Specifically, in order to restrict the latent topics to the label set of each document in CLLDA, we define an indicator function $I^{(d)}(k)$ as follows:  
\begin{eqnarray}
I^{(d)}(k)=
\begin{cases}
1& \text{ if the $k^{th}$ topic is in the label set of}\\
 & \text{ the $d^{th}$ document.}\\
0& \text{ otherwise. }
\end{cases}
\end{eqnarray}

If the word $w_{d,i}$ belongs to some concepts in the given concept knowledge base, the sampling probability for a topic and a concept in position $i$ in document $d$ can be expressed as follows:
\begin{equation}
\small
\begin{aligned}
&P(z_{d,i}=k,c_{d,i}=j|\textbf{w},\textbf{e}_{-(d,i)},\textbf{z}_{-(d,i)};\alpha,\beta)\propto I^{(d)}(k) \\
&\cdot\frac{\beta_{k,c_{d,i}}+n^{(c_{d,i})}_{-(d,i),k}}{\sum^{E}_{x=1}\beta_{k,x}+n^{(\cdot)}_{-(d,i),k}}\cdot
\frac{\alpha_{k}+n^{(d)}_{-(d,i),k}}{\sum^{K}_{t=1}\alpha_{t}+n^{(d)}_{-(d,i),\cdot}}\cdot P(w_{d,i}|c_{d,i})
\end{aligned}
\end{equation}

And if the word $w_{d,i}$ does not belong to any concept in the given concept knowledge base, the sampling probability for a topic in position $i$ in document $d$ can be expressed as follows:
\begin{equation}
\begin{aligned}
&P(z_{(d,i)}=k|\textbf{w},\textbf{e}_{-(d,i)},\textbf{z}_{-(d,i)};\alpha,\beta)\propto I^{(d)}(k) \\
&\cdot\frac{\beta_{k,e_{d,i}}+n^{(e_{d,i})}_{-(d,i),k}}{\sum^{E}_{x=1}\beta_{k,x}+n^{(\cdot)}_{-(d,i),k}}\cdot
\frac{\alpha_{k}+n^{(d)}_{-(d,i),k}}{\sum^{K}_{t=1}\alpha_{t}+n^{(d)}_{-(d,i),\cdot}}
\end{aligned}
\end{equation}
where $I^{(d)}(k)$ is the indicator function, and other notations have the same meaning as that in CLDA stated above.
  
Finally, the parameter can be estimated as follows:
\begin{equation}
\hat{\phi}_{e,k}=
\frac{\beta_{k,e}+n^{(e)}_{\cdot,k}}{\sum^{E}_{x=1}\beta_{k,x}+n^{(\cdot)}_{\cdot,k}}
\end{equation}
\begin{equation}
\begin{aligned}
\hat{\theta}_{k}^{(d)}=\frac{\alpha_{k}+n^{(d)}_{\cdot,k}}{\sum^{K}_{t=1}\alpha_{t}+n^{(d)}_{\cdot,\cdot}}
\end{aligned}
\end{equation}
the notations have the same meaning as that in CLDA stated above.  

\section{Experiment}
\subsection{Experiment Setting}
We conducted the experiments on two real datasets. One of them, called \textbf{Conf}, contains 2,317 full papers of four conferences ( CIKM, SIGIR, SIGKDD and WWW ) of three years (2011 $\sim$ 2013). And the other dataset named \textbf{AP} is a public dataset, which contains more than 106K full Associated Press news articles published in 1989. Both of the raw datasets contain more than 2 million concepts according to Probase, which is much larger than the size of vocabulary. It leads to the imbalance between concepts size and vocabulary size.    
Moreover, lots of concepts in Probase Concept Graph are similar to each other associated with the same word. For example, the word \emph{microsoft} associates with concept \emph{company}, concept \emph{software company} and concept \emph{technology company}, which are semantically similar. In order to address this issue, we use the concept clustering results provided by Probase, to reduce the number of concepts. Totally, it contains 4,819 concept clusters. In the above example, all concepts about \emph{company} can be represented by a concept cluster \emph{company}.  

\begin{table}[tb!]
\centering
\caption{The statistics of the datasets.}
\begin{tabular}{|c|c|c|}
\hline
Datasets & \textbf{Conf} & \textbf{AP} \\
\hline
Size of Documents & 2317 & 106222 \\
\hline
Size of Concepts & 4740 & 4773 \\
\hline
Size of Vocabulary & 18487 & 38419 \\
\hline
\end{tabular}
\end{table}

The statistics of the two datasets are summarized in Table 1. And we conduct all the experiments on a server with an Intel(R) Xeon(R) CPU E5-2683 v3 @ 2.00GHz and 125GB memory. In the rest, we will compare the proposed models with corresponding baselines in terms of case study and perplexity. 
 For all models, we set the number of iterations in each collapsed Gibbs sampler as 1000,   
and set the same initial hyperparameters, where $\alpha$ and $\beta$ both equal to 0.01.  

\subsection{Experiments for CLDA}

In the experiments, we removed the standard stop words for both datasets, and then we further removed words that occurred less than ten times. We trained the two topic models and set the number of topics as 100.

\subsubsection{Case Studies}

Table 2 and Table 3 show top ten words and concepts associated with 5 topics learned on \textbf{Conf} and \textbf{AP} respectively. The topics learned by CLDA were matched to a LDA topic with smallest Kullback-Leibler divergence. It is noted that the bold phrase in the third column of Table 2 and Table 3 is the clustering concepts where pattern ``A, ..., B; C'' means ``A, ..., B'' is similar to ``A, ..., C'', and the un-bold   word is the atomic concepts.  
\begin{table*}[htb!]\footnotesize
\caption{Top ten words and concepts associated with five topics learned on \textbf{Conf}.}
\begin{tabular}{|m{0.2\columnwidth}<{\centering}|m{0.5\columnwidth}<{\centering}|m{0.65\columnwidth}<{\centering}|m{0.5\columnwidth}<{\centering}|}
\hline\noalign{\smallskip}
\tiny{\backslashbox {Topic}{Term}} & Words in LDA & Concepts in CLDA & Words in CLDA \\
\noalign{\smallskip}
\hline
\noalign{\smallskip}
Topic 1 &
similar, pair, weight, measure, compute, approach, vector, base, define, compare &
\textbf{relaxation method;technique}, \textbf{material}, \textbf{activity}, partite, \textbf{operation}, optime, \textbf{company}, \textbf{label}, solute, greedy&
lempel, constrain, entity-relationship, superset, a-priori, in-situ, nearestneighbor, k-nn, z-score, dkt \\
\hline
Topic 2 &
trust, signature, layer, fingerprint, propage, base, bob, similar, credible, result &
attribute, \textbf{product}, \textbf{device}, \textbf{movie;film}, \textbf{activity}, \textbf{company}, \textbf{relaxation method;technique}, \textbf{covariates;question}, numer, category&
attribute, evict, isol, attend, renew, termset, tation, hjk, bioport, vaio\\
\hline
Topic 3 &
feature, predict, perform, set, number, model, regress, problem, inform, baseline &
propose, perform, baseline, denote, \textbf{protein}, experi, calcul, equate, introduce, specify&
propose, perform, baseline, denote, experi, calcul, equate, introduce, specify, outperform\\
\hline
Topic 4 &
select, approach, result, strategy, set, evalu, base, perform, combine, quality &
\textbf{handtools;hand tool}, \textbf{top brand name;brand}, google, provide, specify, \textbf{protein}, \textbf{artist}, explore, \textbf{work}, \textbf{leader}&
t-test, waste, bubble, taste, recipe, dsat, logy, iew, geng, klout \\
\hline
Topic 5 &
role, subgroup, equival, reachable, vertic, degree, snapshot, vertex, approach, show &
compute, \textbf{product}, effic, \textbf{device}, memory, top-k, \textbf{movie;film}, \textbf{covariates;question}, large, \textbf{datatypes;simple variable}&
compute, effici, evict, isol, attend, renew, termset, tation, hjk, bioport \\
\hline
\end{tabular}
\end{table*}

\begin{table*}[htb!]\footnotesize
\caption{Top ten words and concepts associated with five topics learned on \textbf{AP}.}
\begin{tabular}{|m{0.2\columnwidth}<{\centering}|m{0.5\columnwidth}<{\centering}|m{0.65\columnwidth}<{\centering}|m{0.5\columnwidth}<{\centering}|}
\hline\noalign{\smallskip}
\tiny{\backslashbox {Topic}{Term}} & Words in LDA & Concepts in CLDA & Words in CLDA \\
\noalign{\smallskip}
\hline
\noalign{\smallskip}
Topic 1 &
parti, govern, elect, polite, leader, opposit, member, power, parliament, democrat &
\textbf{benefit}, \textbf{covariates;question}, spend, income, \textbf{book}, increase, \textbf{critic}, \textbf{liquid}, \textbf{apple;cultivars}, \textbf{limitation}&
effortless, carnal, pat, intellect, misdemeanor, asylum, laundromat, proud, hispanic-american, ivorian \\
\hline
Topic 2 &
bush, preside, reagan, white, house, administr, fitzwat, nation, quayl, secretary &
committee, senate, \textbf{facility factor;feasibility factor}, \textbf{writer;author}, \textbf{staff;job}, legisl, \textbf{top brand name;brand}, republican, approve, congression&
committee, senate, legisl, jule, mahdi, balcony, guin, ghali, rhetor, vorontsov\\
\hline
Topic 3 &
agreement, talk, negoti, agree, propose, plan, reach, meet, sign, side &
\textbf{noise sensitive use;setting}, \textbf{activity}, \textbf{personnel action}, began, early, begin, decade, large, remain, slowly&
unfriend, abul, noose, all-pro, shipboard, hamburg, edge, moscow, winooski, scatter \\
\hline
Topic 4 &
company, busy, opere, corp, product, service, amp, market, industry, firm &
\textbf{color}, \textbf{stones}, govern, polite, \textbf{instrument}, opposite, minist, communist, parliament, preside&
olive-green, langan, vidalia, yellowish, blue-green, parti, blond, govern, reddish, crimson\\
\hline
Topic 5 &
million, year, billion, percent, share, earn, quarter, sale, total, profit &
\textbf{device}, \textbf{product}, \textbf{covariates;question}, \textbf{car}, vehicle, \textbf{accessory}, \textbf{site-specific information}, automak, made, assemble&
destruct, imagine, lewi, hurl, seat-belt, deadpan, passageway, vinson, c'mon, blinder\\
\hline
\end{tabular}
\end{table*}

From the two tables, we can see that a topic can be reflected by concepts, and a topic can be represented by a distribution of concepts. For example, according to the top words for topic 4 in Table 2 learned by LDA, its topic may be \emph{trade}. In the same topic, there are many concepts learned by CLDA  correlating with that topic, such as \emph{handtools;hand tool}, \emph{top brand name;brand}, \emph{provide} and \emph{specify}. Meanwhile, many words learned by CLDA in the same topic are related to these concepts, such as \emph{t-test}, \emph{waste}, \emph{bubble}, \emph{taste} and \emph{recipe}.  

Another example, as for the top ten words for topic 2 in Table 3 learned by LDA, the topic may be \emph{politics}. In the same topic, concepts, such as \emph{committee}, \emph{senate}, \emph{facility factor;feasibility factor}, \emph{legisl} and republican learned by CLDA are related to topic \emph{politics}, and words learned by CLDA in the same topic are related to these concepts.

Therefore, our proposed CLDA performs better than LDA, and thus our conceptualization method for topic modeling sounds good.

\subsubsection{Perplexity}


\begin{table*}[t!]\footnotesize
\caption{Top ten words and concepts associated with five topics learned on \textbf{Conf}.}
\begin{tabular}{|m{0.2\columnwidth}<{\centering}|m{0.45\columnwidth}<{\centering}|m{0.75\columnwidth}<{\centering}|m{0.45\columnwidth}<{\centering}|}
\hline\noalign{\smallskip}
\tiny{\backslashbox {Label}{Term}} & Words in LLDA & Concepts in CLLDA & Words in CLLDA \\
\noalign{\smallskip}
\hline
\noalign{\smallskip}
social networks &
social, network, number, model, graph, set, inform, result, show, problem &
\textbf{socioeconomic variable}, \textbf{limitation} , \textbf{exercise program;class}, \textbf{operation}, \textbf{asset}, \textbf{activity}, \textbf{perturbation}, \textbf{product}, \textbf{basic contact information;contact information}, \textbf{anomaly} &
network, social, user, number, result, set, inform, node, topic, figure \\
\hline
wikipedia &
wikipedia, article, inform, feature, test, set, word, page, entity, data &
\textbf{activity}, \textbf{poem}, article, \textbf{exercise program;class}, \textbf{limitation}, \textbf{answer}, \textbf{metric}, \textbf{construct}, \textbf{requirement}, \textbf{disadvantage} &
wikipedia, article, feature, inform, entity, evalu, page, category, figure, table\\
\hline
query recommender systems &
query, term, suggest, recommend, node, compute, model, set, result, graph &
query, \textbf{limitation}, \textbf{artifact}, \textbf{writing system;script}, suggest, \textbf{activity}, \textbf{reinforcers;essential}, \textbf{requirement}, \textbf{famous name}, \textbf{high wear area} &
query, suggest, recommend, model, user, generate, set, list, node, qfg\\
\hline
social media &
social, user, media, inform, data, number, work, time, figure, twitter &
\textbf{limitation}, \textbf{exercise program;class}, \textbf{activity}, \textbf{asset}, \textbf{tax implication}, \textbf{operation}, \textbf{construct}, \textbf{company}, \textbf{perturbation}, \textbf{computer function;complex function}&
social, user, media, inform, topic, number, result, network, set, work\\
\hline
data mining &
data, mine, set, inform, result, system, work, number, provide, perform &
\textbf{activity}, \textbf{limitation}, \textbf{requirement}, \textbf{skill tab}, \textbf{asset}, \textbf{metric}, \textbf{mega-projects}, \textbf{operation}, \textbf{construct}, \textbf{reinforcers;essential}&
data, mine, result, set, user, inform, learn, model, provide, number\\
\hline
\end{tabular}
\end{table*}

\begin{figure}[tb!]
\setlength{\abovecaptionskip}{-5pt}
\setlength{\belowcaptionskip}{-18pt}
\begin{minipage}{1.0\linewidth}
    \begin{center}
        \begin{tabular}{m{0.48\linewidth}m{0.48\linewidth}} \subfloat[\textbf{Conf}]{\includegraphics[width=0.98\linewidth]{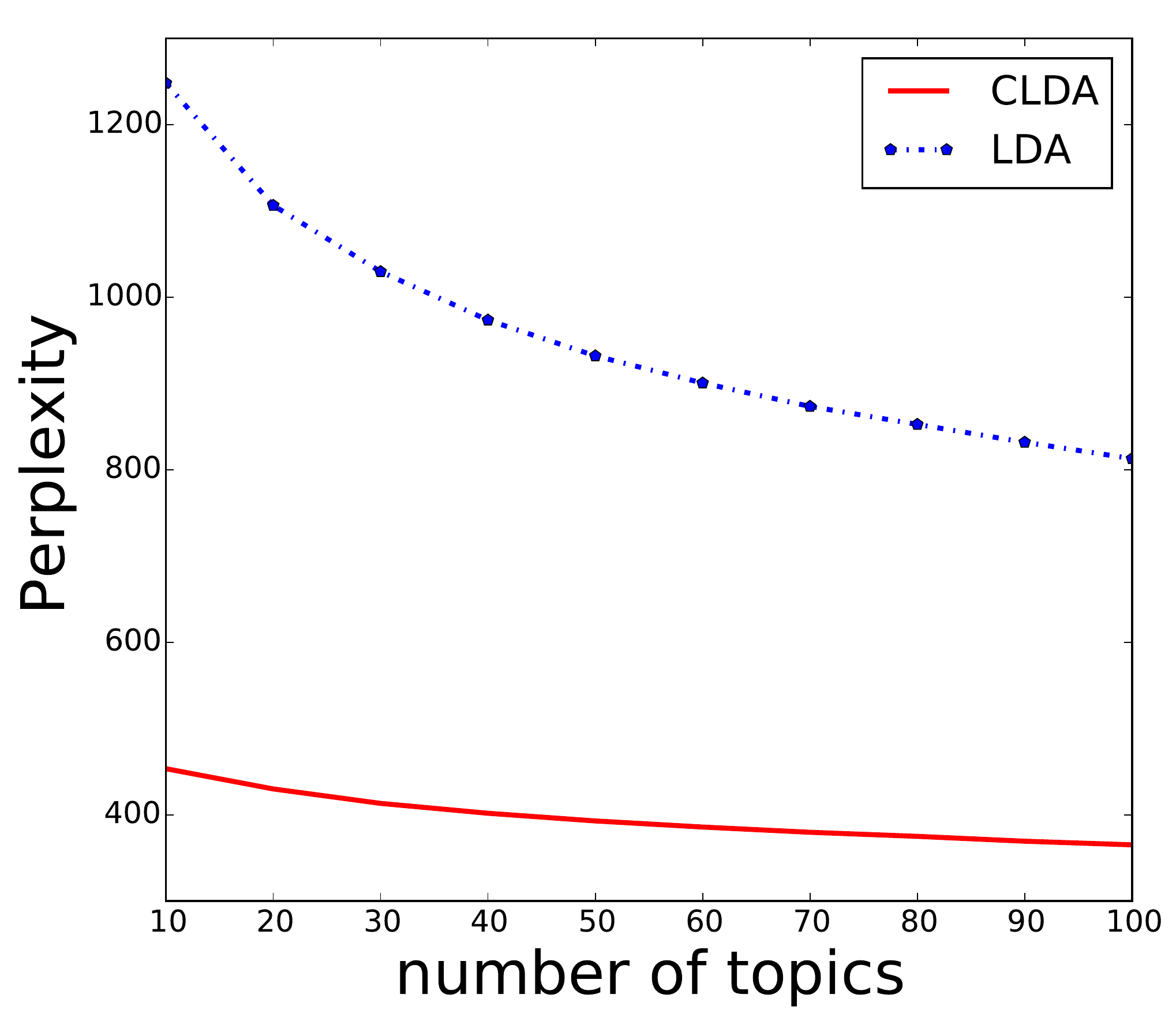}}&
        \subfloat[\textbf{AP}]{\includegraphics[width=0.98\linewidth]{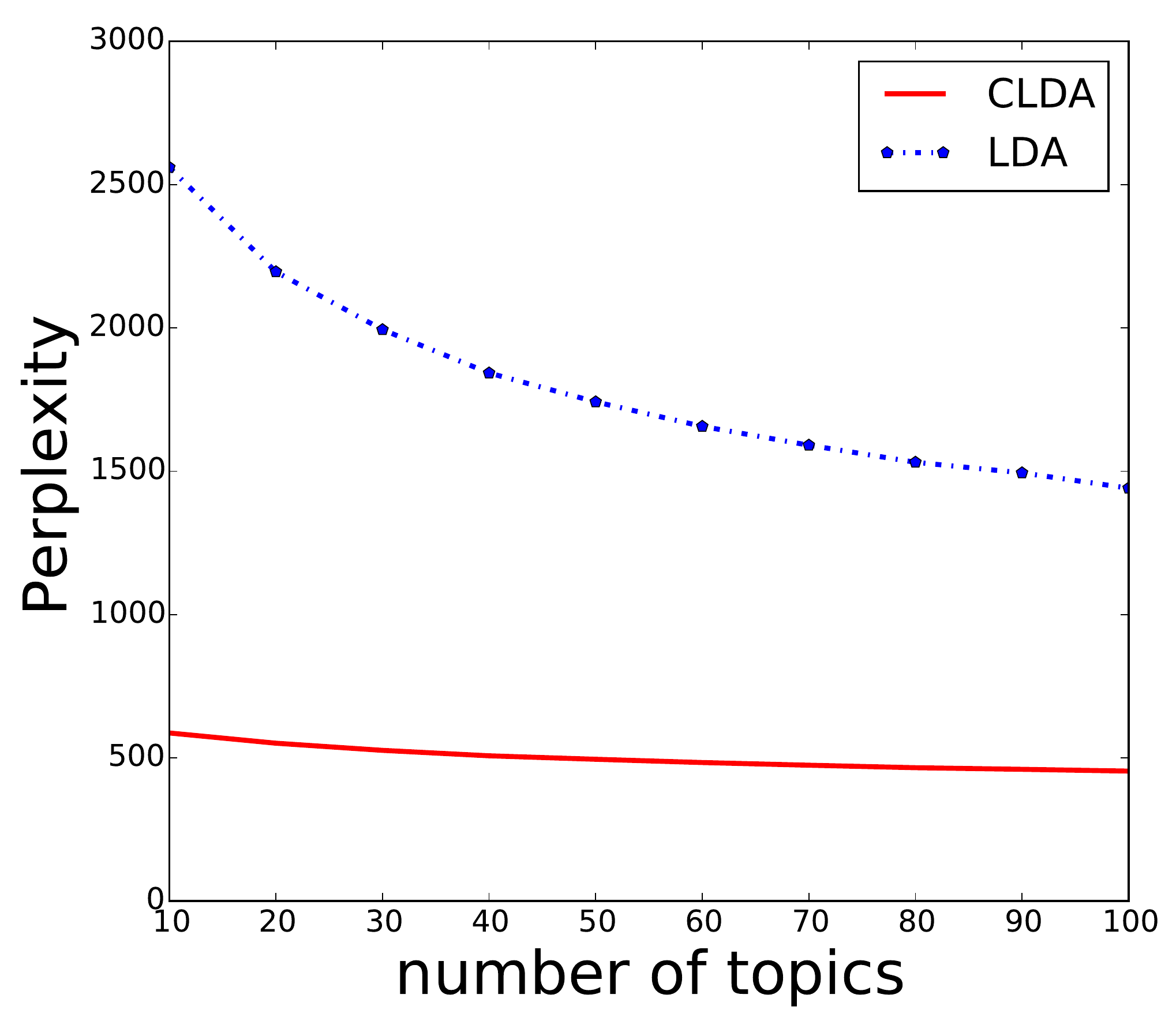}}
        \end{tabular}
    \end{center}
\caption{A comparison of the perplexity of CLDA and LDA with different number of topics on two datasets. }
\label{fig:geo_distribution}
\end{minipage}
\end{figure}

In this experiment, we trained both CLDA and LDA for ten times with different number of topics varying from 10 to 100 in turn. We computed the perplexity of the proposed CLDA and LDA, which can quantificationally measure the quality of different models. A lower perplexity score indicates better generalization performance. The perplexity can be computed as follows:
\begin{equation}
\emph{perplexity}=exp{\{-\frac{\Sigma^{M}_{d=1}\log \emph{p}(\textbf{w}_{d})}{\Sigma^{M}_{d=1}N_{d}}\}}
\end{equation}
Figure 3 shows the perplexity of the two models on the two datasets for different number of topics. As we can see in Figure 3, the perplexity curves decrease as the number of topics increases on both of the two datasets. The perplexity value of CLDA is much smaller than that of LDA, which indicates that CLDA performs significantly better than LDA.

\subsection{Experiments for CLLDA}

In this experiment, we train CLLDA and LLDA over \textbf{Conf} dataset. The keywords of each paper will be used as labels of the corresponding  paper, and the number of labels for the whole dataset is 4760. 

\subsubsection{Case Study}

Table 4 shows top ten words and concepts from five topics learned on \textbf{Conf}, where we can easily acquire the concepts under topics. From Table 4, we can learn that concepts is very important in understanding a document. For example, in topic \emph{social media}, concepts such as \emph{limitation}, \emph{activity} and \emph{computer function;complex function}, are different aspects of the topic \emph{social media}, and the words like \emph{social}, \emph{user}, \emph{media} are related to these concepts.
Therefore, from the results we know that the proposed model performs better than the baseline, which means our conceptualization method for topic modeling sounds good, and our assumption is more reasonable.  


\subsubsection{Perplexity}
To compute perplexity for the two models with different number of topics, we segment documents in \textbf{Conf} into ten groups, and the first nine groups all contain 200 documents, while the last group contains the rest of the documents. We train CLLDA and LLDA for ten times. The first time we use the first group, the second we use the first two groups, and so on. The comparison of the two models' perplexity is shown in Figure 4. Since the data for training with different number of topics is different, the monotonic trend of the same curve is not comparable. Here, we only compare the perplexity values of the two curves under the same number of topics. It is apparent that CLLDA performs much better than the original LLDA.

\begin{figure}[t]
\centering
 \includegraphics[width=0.65\linewidth]{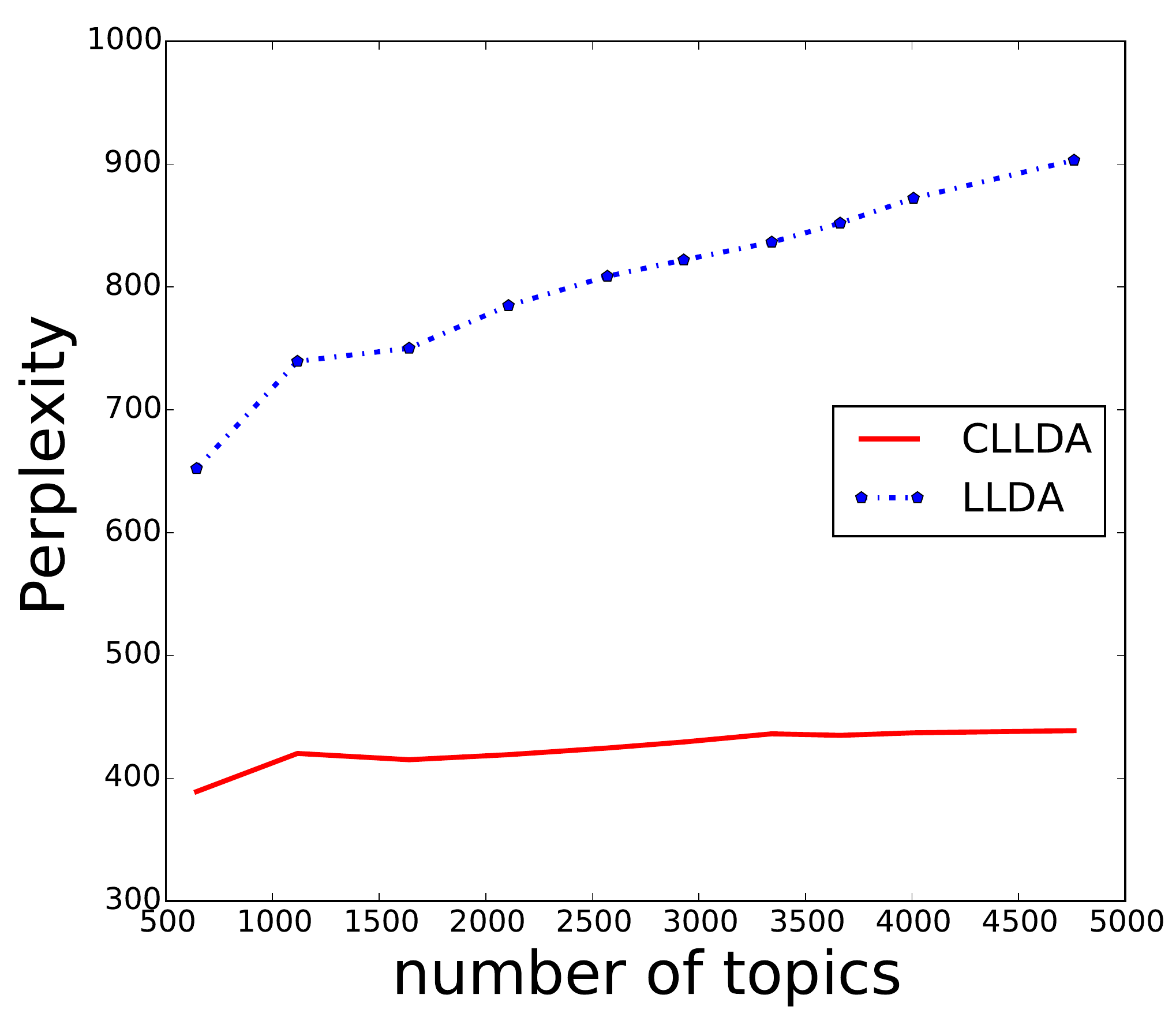}
\caption{A comparison of the perplexity of CLLDA and LLDA with different number of topics on \textbf{Conf}.}  
\label{fig:geo_distribution_cllda}
\end{figure}

\section{Conclusion and Future Work}
In this paper, we propose a novel assumption of four-layer hierarchical Bayesian structure for topic modeling, which adds a latent concept layer between topic layer and word layer in the traditional assumption. To verify the effectiveness of the novel assumption, we apply the assumption in two representative topic models (LDA and LLDA).   
Extensive experiments have been conducted on two real datasets.   
The experimental results show that the proposed assumption performs better than the traditional assumption.  

In the future, we will further verify the novel four-layer assumption in more topic models over more datasets.  Also, we will explore the use of the new assumption in multimedia data, such as images and videos.    



\bibliographystyle{named}
\bibliography{ctm}

\end{document}